\documentclass{article}

\usepackage[nonatbib,final]{neurips_2024}

\usepackage[utf8]{inputenc} %
\usepackage[T1]{fontenc}    %
\usepackage{hyperref}       %
\usepackage{url}            %
\usepackage{booktabs}       %
\usepackage{amsfonts}       %
\usepackage{nicefrac}       %
\usepackage{microtype}      %
\usepackage{xcolor}         %
\usepackage{amsmath}
\usepackage{graphicx}
\usepackage{float}

\usepackage[sorting=none]{biblatex}

\title{Investigating Sensitive Directions in GPT-2: An Improved Baseline and Comparative Analysis of SAEs}

\author{%
  Daniel J. Lee\footnotemark[1] \\
  Department of Biomedical Informatics\\
  Harvard Medical School\\
  Boston, MA \\
  \And
  Stefan Heimersheim \\
  Apollo Research \\
  London, UK \\
}

\begin{document}

\maketitle
\def\thefootnote{*}\footnotetext{\href{mailto:daniel_lee@g.harvard.edu}{daniel\_lee@g.harvard.edu}}

\begin{abstract}
    Sensitive directions experiments attempt to understand the  computational features of Language Models (LMs) by measuring how much the next token prediction probabilities change by perturbing activations along specific directions. We extend the sensitive directions work by introducing an improved baseline for perturbation directions. We demonstrate that KL divergence for Sparse Autoencoder (SAE) reconstruction errors are no longer  pathologically high compared to the improved baseline. We also show that feature directions uncovered by SAEs have varying impacts on model outputs depending on the SAE's sparsity, with lower \(L0\) SAE feature directions exerting a greater influence. Additionally, we find that end-to-end SAE features do not exhibit stronger effects on model outputs compared to traditional SAEs.
\end{abstract}

\section{Introduction}

One of the primary goals of mechanistic interpretability is to uncover the variables that neural networks use in their computation. This task is complicated by polysemanticity, a phenomenon where a single neuron activates in response to multiple seemingly unrelated features \cite{Arora2016-cq, Olah2017-dd}. Recent studies \cite{Bricken2023-ay, Cunningham2023-yl} have employed an unsupervised dictionary learning algorithm called Sparse Autoencoders (SAEs) to disentangle LM activations into sparse, linear combinations of feature directions. While SAEs show significant promise \cite{Templeton2024-dw}, there is limited dataset-independent evidence that the features found by SAEs are indeed true abstractions used by the LMs.

Several works have sought to understand these abstractions by observing how much the next token prediction probabilities change when activations are perturbed along certain directions, a technique hereinafter referred to as sensitive direction analysis. \textcite{Heimersheim2024-vj} demonstrated, for example, that perturbing from one real activation towards another real activation changes the model output earlier (shorter perturbation lengths) than perturbing into random directions. This finding supports the hypothesis that perturbations along true feature directions have a greater impact on model outputs compared to other directions, motivated by toy models of computation in superposition \cite{Hanni2024-oq}.

Sensitive direction analyses have been also used to evaluate Sparse Autoencoders (SAEs). Perturbations along the SAE feature directions appear to alter the model output more than random directions, suggesting that SAEs successfully uncover important “levers” used by the model \cite{Lindsey2024-zn}. However, SAE-reconstructed activation vectors also alter the model output much more than random perturbations of the same L2 distance from the base activation, an observation that puzzled the interpretability community \cite{Gurnee2024-mm}. This phenomenon was characterized as a pathological behavior of SAE reconstruction errors.

\paragraph{Our contribution} In this paper, we expand on the sensitive directions work. We show that:

\begin{itemize}
  \item \textcite{Heimersheim2024-vj}’s sensitive direction baselines were flawed in that the perturbation direction involved subtracting the original activation. We propose an improved baseline direction (called \textit{cov-random mixture}) which does not use the original activation.
  \item \textcite{Gurnee2024-mm}’s KL-divergence for SAE reconstruction errors no longer seems pathologically high when we use this improved baseline.
  \item Perturbations into SAE feature directions reveal that (1) SAE directions have smaller or greater impact on the model output than our baseline, depending on the SAE type and \(L0\), and (2) lower \(L0\) SAE feature directions have a greater impact on the model output than higher \(L0\) SAE feature directions.
  \item Feature directions from end-to-end SAEs do not exhibit a greater influence on the model output than those from traditional SAEs.
\end{itemize}

\section{Experimental Methods}

The experiments described in this report focus on perturbing an activation within the residual stream of GPT2-small. Specifically, we perform perturbation as follows:
\[
x \leftarrow x^{\text{base}} + \alpha d
\]
where \(x^{\text{base}}\) represents the original activation, \(\alpha\) is the perturbation length, and \(d\) is a unit direction vector. To assess the impact on the model's output, we use the KL divergence of the next token prediction probabilities (more specifically, KL(original prediction | prediction with substitution)). Unless if otherwise stated, the perturbations are applied in Layer 6 resid\_pre. Layer 6 was chosen because \textcite{Braun2024-fl}’s main results focus on end-to-end SAEs on Layer 6 activations.

\paragraph{Data} The experiments are performed on approximately 2 million tokens (16,000 sequences, each with a length of 128) from Openwebtext. We also run a subset of the experiments (Figure 1) for 15 million tokens from Openwebtext, and confirm that the result stays the same (see Appendix B). We perturb activations for all token positions.

\paragraph{Extrapolation} When we extrapolate the perturbation vector, we extend the vector from length 0 to 101 (the mean L2 distance between two actual activations in Layer 6 resid\_pre is 81.59). Our results mainly focus on the resulting curves of KL vs perturbation length or L2 distance at Layer 11 vs perturbation length. We use the mean of KL or mean of L2 across the 2 million tokens as our main measure. We use the mean under the assumption that directions with greater functional importance will, on average, induce a more significant change in the model's output.

\section{Developing a Better Baseline}

\textcite{Lindsey2024-zn, Gurnee2024-mm} use random isotropic perturbation as their baseline. Both papers point out that this might be problematic because actual activations are not isotropic, and some sensitivity differences may be explained by that effect. Previous work by \textcite{Heimersheim2024-vj} attempts to address this issue by adjusting the mean and covariance matrix of the randomly generated activations to match real activations. However, the paper's perturbation directions use the direction from the original activation toward another random activation (\(x^{\text{target}} - x^{\text{base}}\)), which includes the negative of the original activation (\(- x^{\text{base}}\)) as a component. This makes it an unfair comparison to directions that do not include the original activation. Therefore, we propose two new baselines (\textit{cov-random mixture} and \textit{real mixture}) where the directions do not include the original activation.

Following is the list of perturbation directions discussed in this section:

\begin{itemize}
  \item \textbf{Isotropic random}: Perturb into a random direction (no subtraction)
  \item \textbf{Cov-random difference}: Perturb along \(d = x^{\text{cov-random}} - x^{\text{base}}\), i.e. from base towards a cov-random activation. This direction was used in \textcite{Heimersheim2024-vj} ("random direction").
  \item \textbf{Cov-random mixture}: Perturb along \(d = x^{\text{cov-random}}_1 - x^{\text{cov-random}}_2\), i.e. into the difference of two randomly generated covariance matrix adjusted activations.
  \item \textbf{Real difference}: Perturb along \(d = x^{\text{real}} - x^{\text{base}}\), i.e. from base towards another real activation. A real activation is sampled from the activations from ~2 million tokens. This direction was used in \textcite{Heimersheim2024-vj} ("random other"). Like the \textit{cov-random difference}, this direction contains the original activation.
  \item \textbf{Real mixture}: Perturb along \(d = x^{\text{real}}_1 - x^{\text{real}}_2\), i.e. into the difference of two real activations (not the original activation). The real activations are sampled from the activations from 2 million tokens. The real mixture no longer contains the original activation.
\end{itemize}

\textit{Mixture} directions are useful baselines because, under the Linear Representation Hypothesis, they are a linear combination of true feature directions (see Appendix C) \cite{Bricken2023-ay}.

\subsection{Comparing different baselines}

\begin{figure}
  \centering
  \includegraphics[width=10cm]{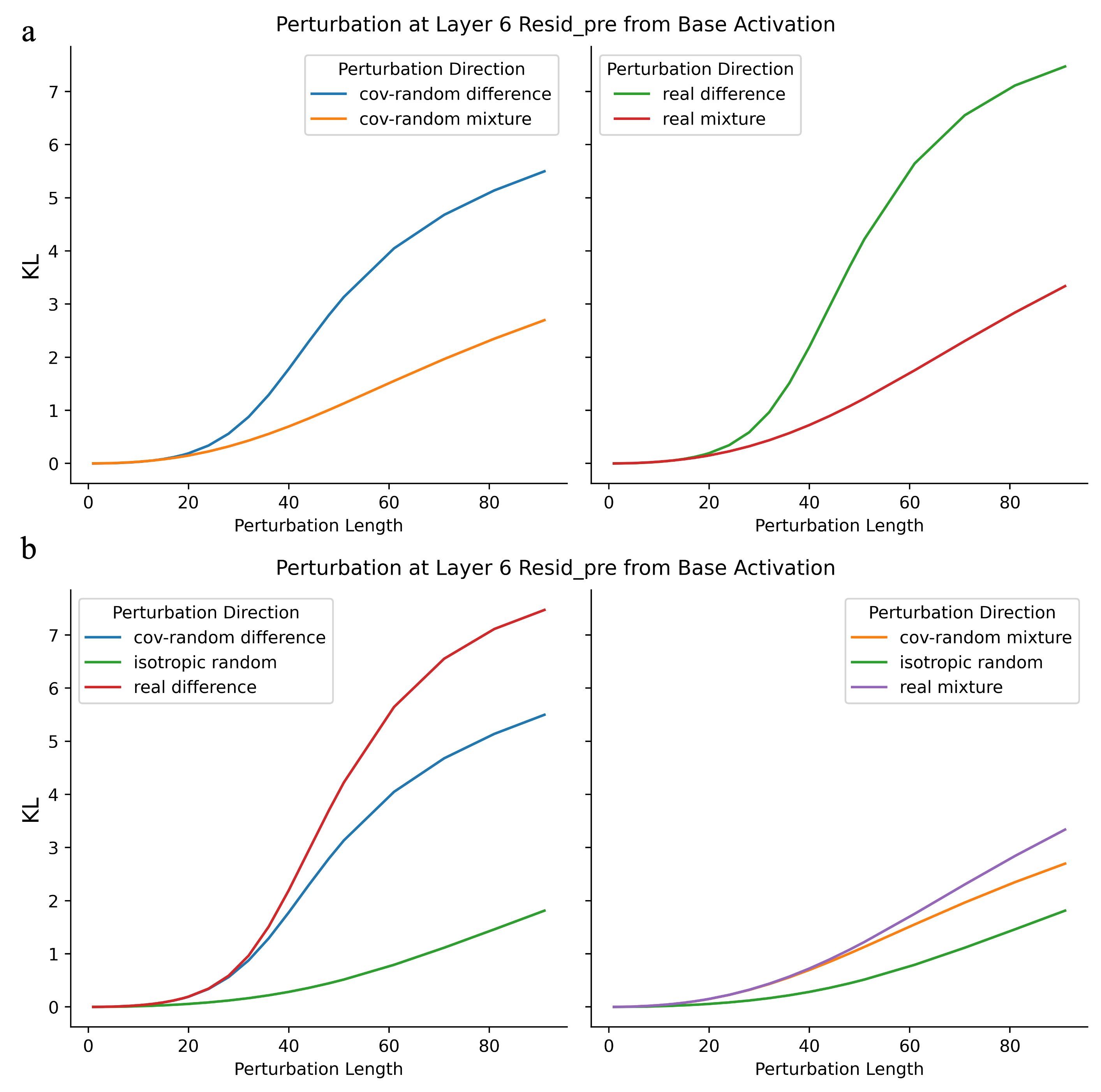}
  \caption{We vary the perturbation length for perturbations in Layer 6 resid\_pre. (a) We compare the \textit{difference} versus \textit{mixture} perturbations. For both \textit{cov-random} (left) and \textit{real} (right) cases, the \textit{difference} perturbations have a greater change in model output than \textit{mixture} perturbations. (b) We compare the \textit{cov-random} versus \textit{real} baselines for \textit{difference} (left) and \textit{mixture} (right) types.}
\end{figure}

On average, perturbation directions that include the negative original activation (\(- x^{\text{base}}\)) cause a greater change in the model output compared to those that do not include the original activation. In Figure 1a, KL for \textit{cov-random difference} is greater than KL for \textit{cov-random mixture} and the KL for \textit{real difference} is greater than KL for \textit{real mixture}. This finding suggests that the \textit{difference} directions may primarily reflect the subtraction of the original activation, which seems related to the observation in \textcite{Lindsey2024-zn} that “feature ablation” has a much greater effect than other perturbations including “feature doubling.” The result supports the use of \textit{mixture} baselines to ensure a fair comparison with directions like SAE features or SAE errors, which do not necessarily involve the original activation.

\textit{Cov-random mixture} directions influence the model's output more significantly than isotropic random directions (see the right plot of Figure 1b). This aligns with the hypothesis that isotropy reduces the influence of perturbations on the model’s logits. Since \textit{cov-random} directions are derived from a multivariate normal distribution and real activations are likely more clustered than normally distributed, \textit{cov-random} directions may not be the ideal baseline. \textcite{Heimersheim2024-vj} observed that \textit{real difference} directions had a stronger effect on the model’s output than \textit{cov-random difference} directions (as shown in the left plot of Figure 1b), which initially suggested that sampling from real activations might provide a better baseline. However, the minimal difference between \textit{real mixture} and \textit{cov-random mixture} directions (shown in right plot of Figure 1b) suggests that \textcite{Heimersheim2024-vj}'s finding was likely influenced by the negative activation component.

\subsection{Revisiting pathological errors under new baselines}

\begin{figure}
  \centering
  \includegraphics[width=12cm]{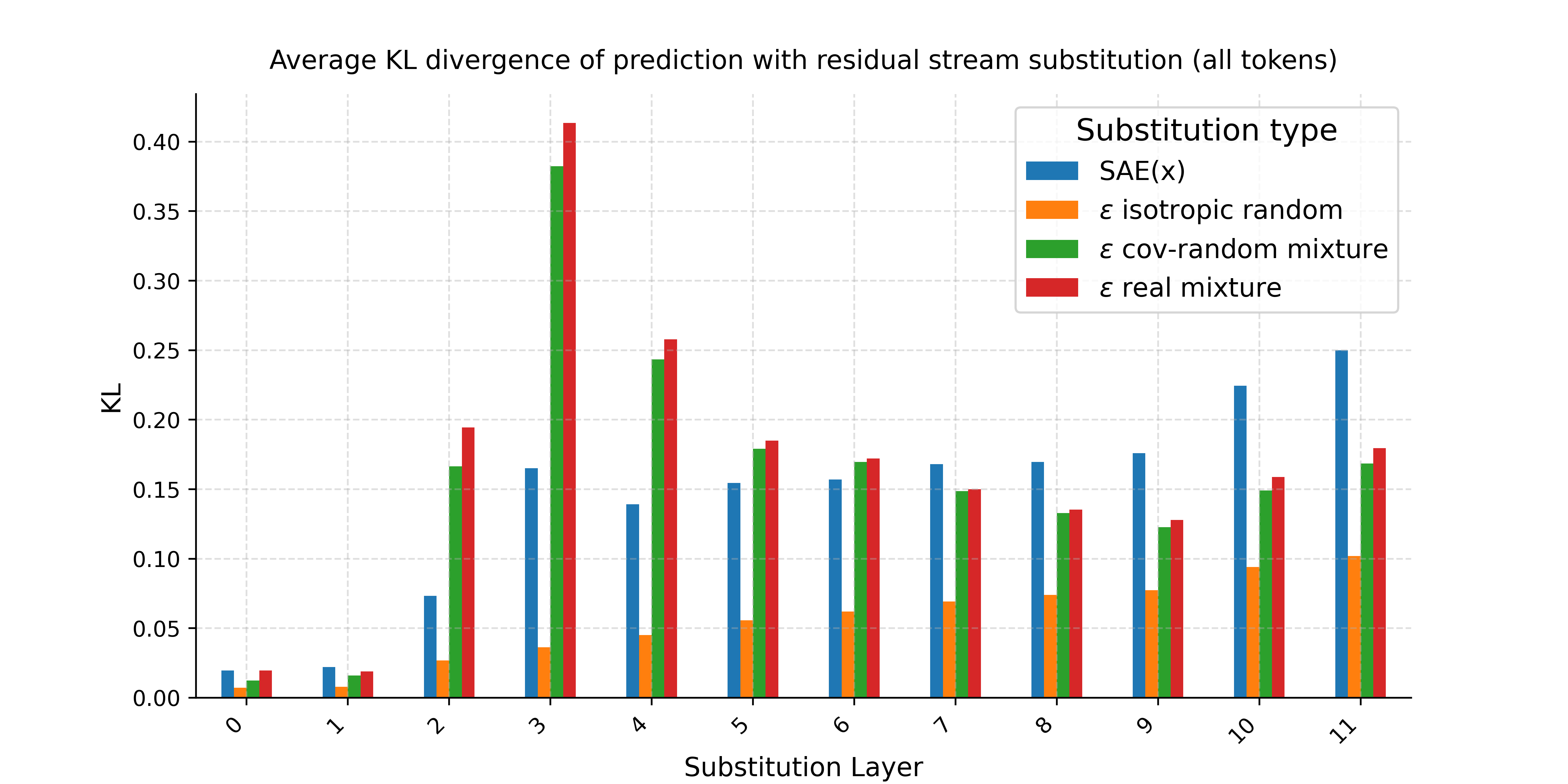}
  \caption{Comparison of the average KL divergence of four different substitution types. On the x-axis we have different GPT2-small layers. SAE from \textcite{Bloom2024-bi} was used.}
\end{figure}

We reran the analysis from \textcite{Gurnee2024-mm}, this time incorporating the two new baselines. We also compared multiple SAEs with different \(L0\) values. Our results confirmed the original finding that substituting the base activation with the SAE reconstruction, SAE(x), changes the next token prediction probabilities significantly more than substituting an isotropically random point at the same distance \(\epsilon\) (Figure 2). When perturbing along the \textit{cov-random mixture} or \textit{real mixture} directions, the average KL divergence is generally closer to that of SAE(x). However, there is considerable variance depending on the layer. For Layer 6, the SAE models across \(L0\) generally seem to have nearly the same KL as that of cov-random mixture (Figure 5). While this suggests that addressing isotropy mitigates the previously observed pathologically high-KL behavior in SAE errors, questions remain about the dependence observed across different layers.

\section{Comparative Analysis of SAEs}

Recently, a new type of SAEs called end-to-end SAEs has been introduced \cite{Braun2024-fl}. End-to-end SAEs aim to identify functionally important features by minimizing the KL divergence between the output logits of the original activations and those of the SAE-reconstructed activations. There are two variants of end-to-end SAEs: e2e SAE and e2e+ds SAE (where ds is short for downstream). \textcite{Braun2024-fl} proposed e2e+ds SAEs as a superior approach because it also minimizes reconstruction errors in subsequent layers (whereas e2e SAEs might follow a different computational path through the network). In this section, we will compare traditional (local) SAEs, e2e SAE, and e2e+ds SAE across various \(L0\)s.

The following are the perturbation directions discussed in this paper:
\begin{itemize}
  \item \textbf{SAE Reconstruction Error Direction}: Perturb along \(d = \text{SAE}(x^{\text{base}})-x^{base}\), i.e. from base activation towards the reconstructed activation.
  \item \textbf{SAE Feature Direction}: Perturb along \(d=W^{\rm dec}_i\), which corresponds to one of the directions \(i\) from the SAE decoder matrix \(W^{\rm dec}\). We choose SAE features that are alive, but not active in the given sequence.
\end{itemize}

\subsection{SAE reconstruction error extrapolation}

\begin{figure}
  \centering
  \includegraphics[width=12cm]{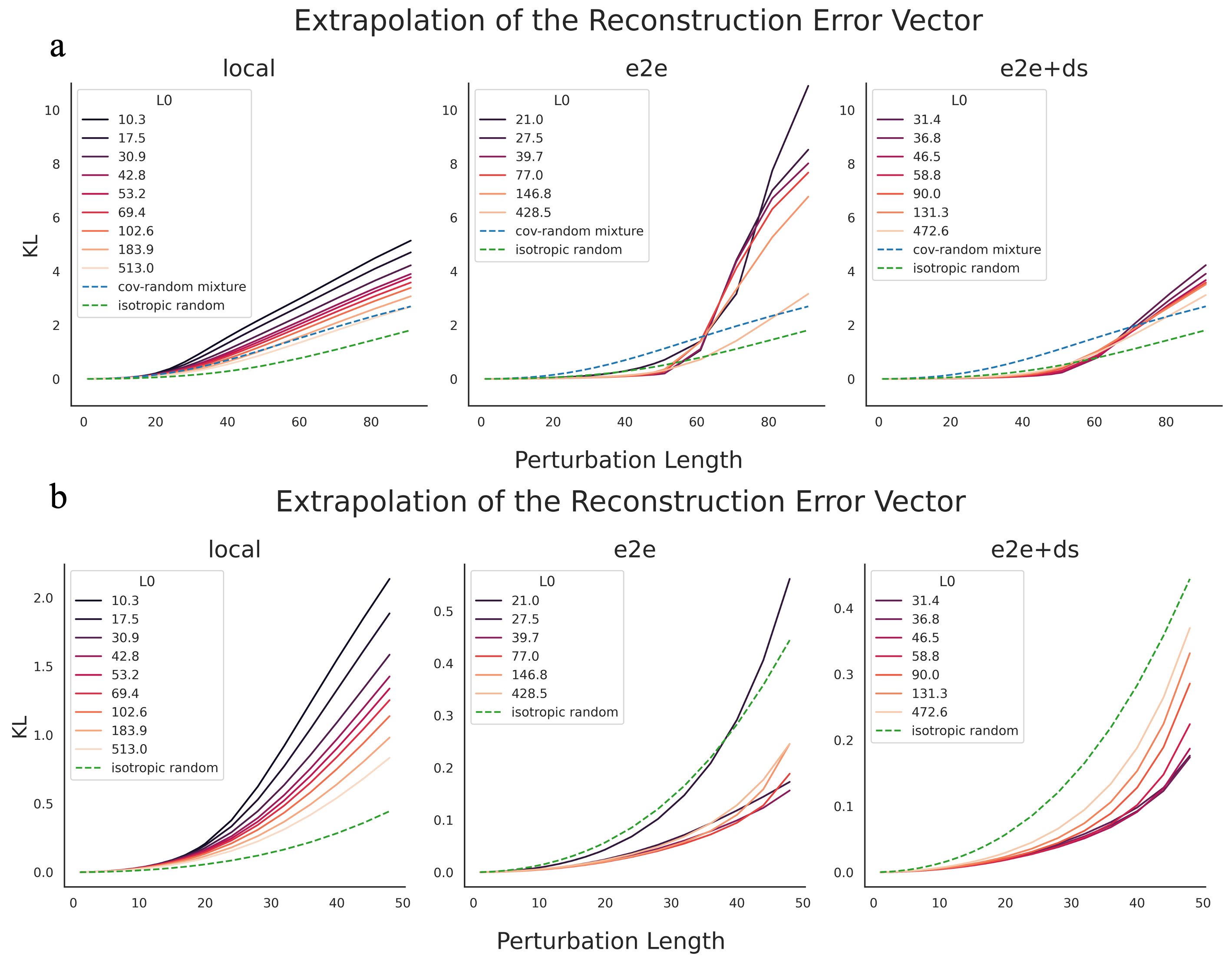}
  \caption{We vary the perturbation length for perturbations in Layer 6 resid\_pre. For each columns we show different SAE model types. We compare the SAE reconstruction error directions with \textit{cov-random mixture} and \textit{isotropic random} directions. We color the lines by different \(L0\) values of the SAEs. (b) is the same as (a), but with a reduced x-axis limit.}
\end{figure}

To gain insight into the model's sensitivity to SAE reconstruction errors, we extend the error directions across various perturbation lengths. The KL curves typically exhibit a consistent pattern: an initial plateau followed by a more linear increase in KL after the plateau. Th plateau aligns with the observation made in \textcite{Heimersheim2024-vj}. For local SAEs, the behavior is straightforward: lower \(L0\) corresponds to a stronger perturbation effect (left plot in Figure 3a). For e2e (and e2e+ds) SAEs, the behavior is more complex: the effect of \(L0\) at small perturbation scales is the opposite of its effect at larger scales. For perturbation lengths below ~50, lower \(L0\) results in greater KL divergence for e2e and e2e+ds SAEs, except for \(L0\) = 21.0 or 27.5 e2e SAEs (middle and right plots in Figure 3b). For perturbations above 70, lower \(L0\) corresponds to a stronger perturbation effect (Figure 3a).

While the curves for the local SAEs are close to the curves for the \textit{cov-random mixture} baseline, the curves deviate a lot for e2e and e2e+ds SAEs. Notably, the curves for e2e and e2e+ds SAEs remain low and then spike up from perturbation length of around 50 (Figure 3a). The former is expected as e2e SAEs generally have a high L2 reconstruction error while having a low KL-divergence).

\subsection{SAE feature extrapolation}

\begin{figure}
  \centering
  \includegraphics[width=12cm]{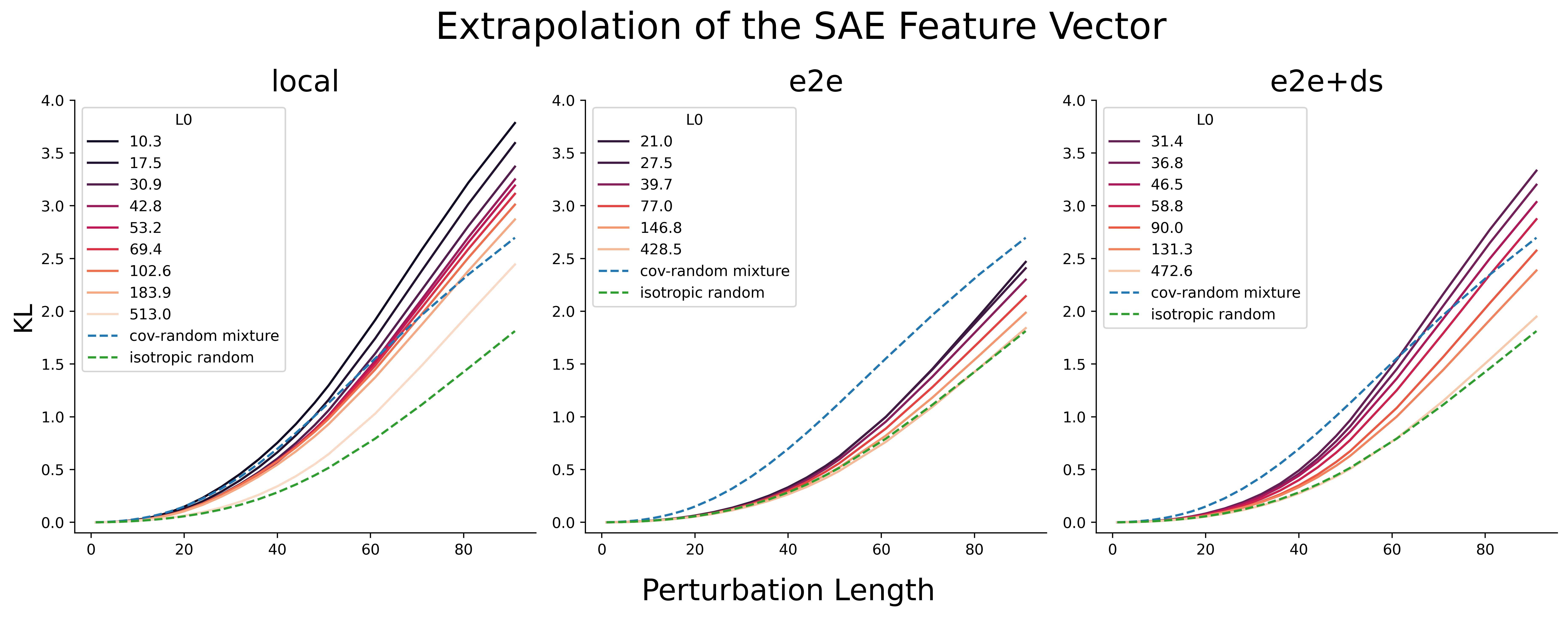}
  \caption{This plot varies the perturbation length for SAE feature directions in Layer 6 resid\_pre. For the three columns, we compare the three different SAE model types . We color the lines by different L0 values of the SAEs.}
\end{figure}

\begin{figure}
  \centering
  \includegraphics[width=12cm]{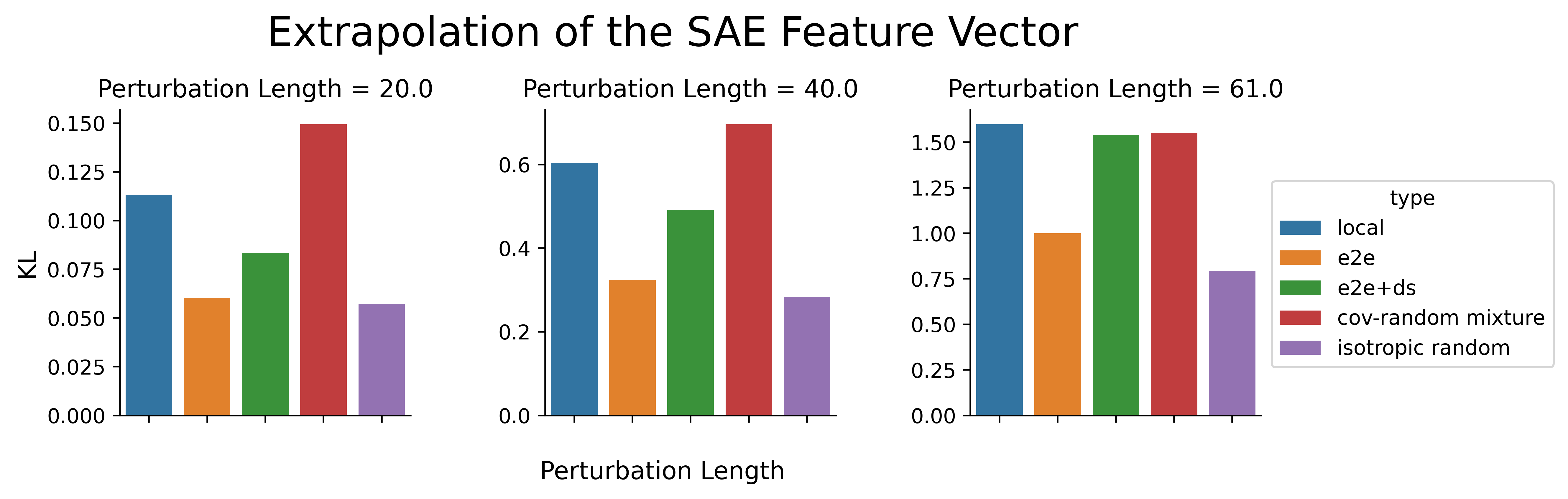}
  \caption{Comparison of the change in model output for various perturbation lengths for different SAE feature directions and baselines in Layer 6 resid\_pre.}
\end{figure}

To explore the functional relevance of SAE features, we extrapolate the SAE feature directions across various perturbation lengths. We select a random SAE feature that is alive, but not active in the given context the token is located in. All SAE features have a greater impact on the model output than \textit{isotropic random} directions (Figure 4). When compared to \textit{cov-random mixture}, the effect varies based on the type of SAE and its \(L0\) value. For all three types of SAEs, lower \(L0\) corresponds to greater change in model output (Figure 4).

We select a specific \(L0\) value to conduct a more detailed comparison of the SAE models (\(L0=30.9\) for local SAE, \(L0=27.5\) for e2e SAE, and \(L0=31.4\) for ds+e2e SAE). Among these, e2e SAE features have the least impact on the model output (Figure 3). At shorter perturbation lengths, local SAE features influence the model more than e2e+ds SAE features, but this difference shrinks as the perturbation lengths increase. We note that using the same \(L0\) may not be a fair way to compare the three SAE models. This is because end-to-end SAEs are known to explain more network performance given the same \(L0\) \cite{Braun2024-fl}.

\paragraph{Discussion} The result was initially surprising because we would have expected that end-to-end SAEs would more directly capture the features most crucial for token predictions. Our hypothesis for the explanation for this observation is that e2e SAE features perform worse because they are more isotropic (see Figure 3(a) from \cite{Braun2024-fl}). This could be an unintended and undesirable consequence of end-to-end SAEs. While e2e SAE might exhibit this behavior, it is unclear to what extent e2e+ds SAE also does this.

\section{Conclusion}
\paragraph{Summary}
In this work, we run sensitive direction experiments for various perturbations on GPT2-small activations. We make several findings. First, SAE errors are no longer pathologically large when compared to more realistic baselines. Second, GPT2-small is more sensitive to lower \(L0\) SAE features. Third, End-to-end SAE features do not exhibit stronger effect on the model output than traditional SAE features.
\paragraph{Limitation}
In this post, we primarily use the mean (of KL) as our main measure. However, relying solely on the mean as a summary statistic might oversimplify the complexity of sensitive directions. For instance, the overall shape of the curve for each perturbation could be another important feature that we may be overlooking. While we did examine some individual curves and observed that real mixture and cov-random mixture generally exhibited greater model output change compared to isotropic random, the pattern was not as clear-cut.

\small
\printbibliography
\normalsize	

\appendix

\section{Additional Figures}

\begin{figure}[!ht]
  \centering
  \includegraphics[width=10cm]{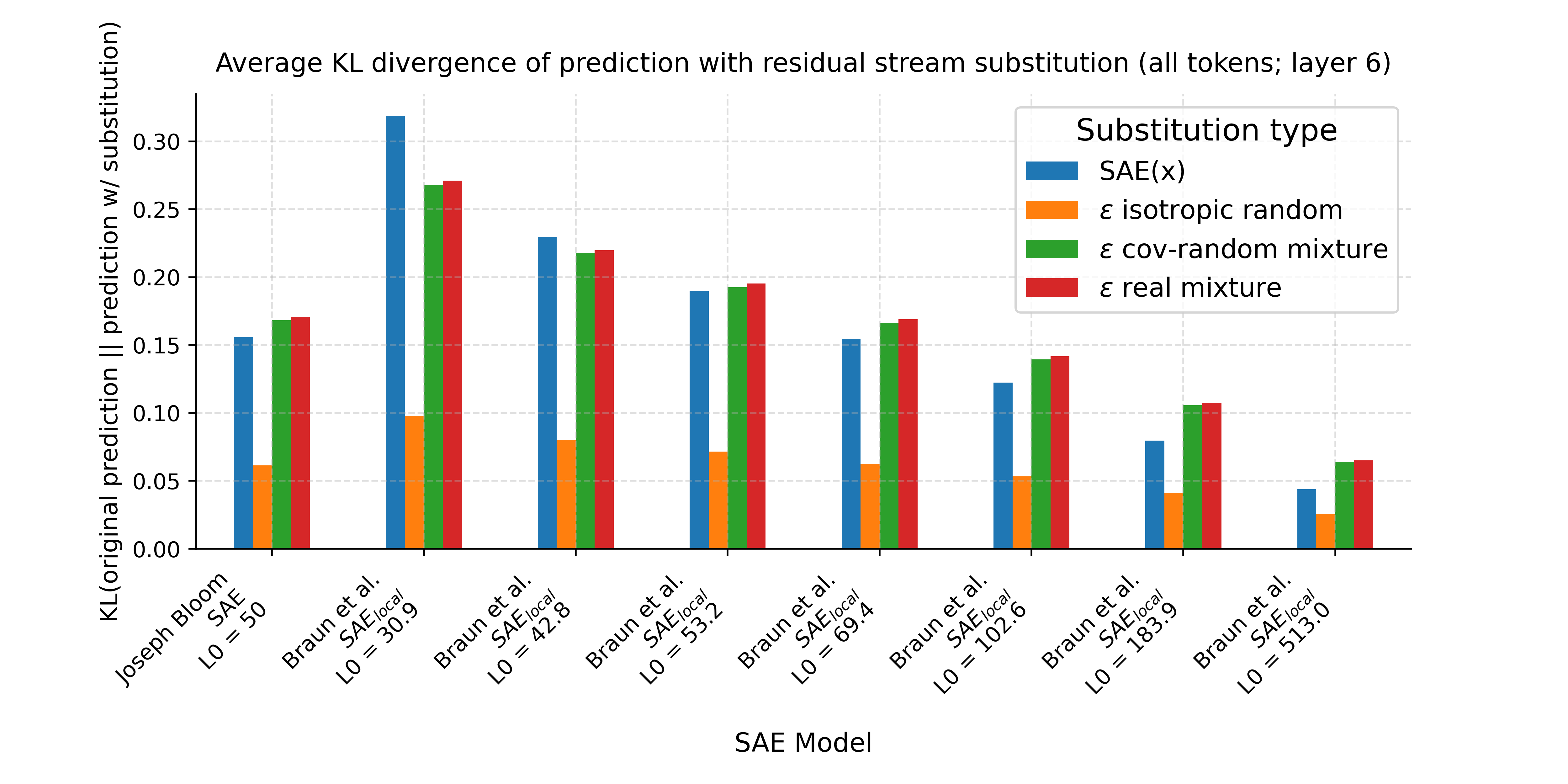}
  \caption{This plot compares the average KL divergence of four different substitution types. On the x-axis we have different SAE models. Joseph Bloom SAE was the SAE used in the original Gurnee 2024 paper. The local SAE from Braun 2024 refers to traditional SAEs. The isotropic random substitutions have a much smaller average KL divergence than other substitution types. Across the various SAE models, the three other substitution types (SAE(x), cov-random mixture, and real mixture) have generally similar average KL divergence.}
\end{figure}
\newpage
\section{Experiments with 15 million tokens}
\begin{figure}[hbt!]
  \centering
  \includegraphics[width=10cm]{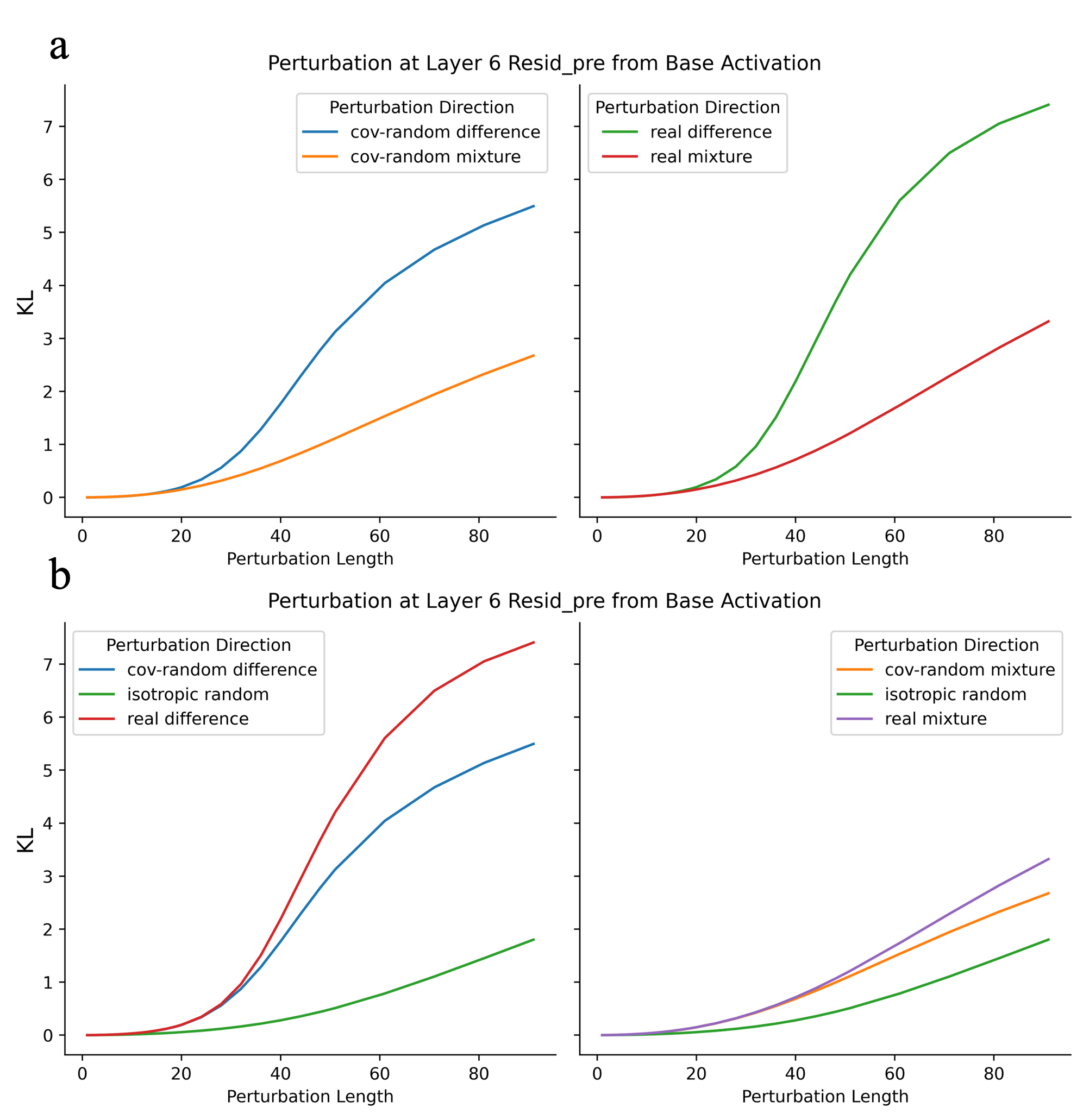}
  \caption{We rerun figure 1 experiment with 15 million tokens.}
\end{figure}

\newpage
\section{Why the Difference Between Two Activations?}

Under the Linear Representation Hypothesis (LRH), we can represent an activation \(x\) as 

\[x \approx b + \sum_i{f_i(x)d_i},\]

where \(f_i(x)\) is the activation of (hypothetical) feature \(i\), \(d_i\) is the unit “direction” vector of feature \(i\), and \(b\) is the bias.

If we take the difference between two activations \(x_1\) and \(x_2\), we get:

\[x_1 - x_2 \approx \sum_i{[f_i(x_1) - f_i(x_2)]d_i}\]

Therefore, assuming LRH, subtracting any two real activations is a linear combination of (hypothetical) true features without the bias term. We note that this will also include “negative features,” which is not expected to be as meaningful in the models.

\end{document}